\documentclass[acmlarge,screen,nonacm]{acmart}
\usepackage{booktabs}
\usepackage{array}
\usepackage{longtable}
\setcopyright{none}
\begin{document}
\title{Component Replacement in LLM Agents: A Critical Scoping Review of Task Benefit and Attribution}
\author{ZHANG Shuyang}
\authornote{Equal contribution; Corresponding author}
\email{25049993g@connect.polyu.hk}
\affiliation{\institution{The Hong Kong Polytechnic University}\city{Hong Kong}\country{China}}
\author{CHANG Jianshuo}
\authornote{Equal contribution}
\affiliation{\institution{The Hong Kong Polytechnic University}\city{Hong Kong}\country{China}}
\renewcommand{\shortauthors}{Zhang and Chang}
\begin{abstract}
{\bf Background:} Choosing an agent controller and deciding which component to improve require different evidence. A replacement can change both the decision rule and what the agent can do after that decision.

{\bf Objectives:} We ask what existing replacement comparisons establish about task benefit and about attribution to local decision quality.

{\bf Methods:} This critical scoping review maps 348 studies and applies a five-step interpretation procedure to 94 comparison and diagnostic records from 42 studies, reading every inconsistent record against its source. Eight purposively selected cases illustrate the procedure.

{\bf Results:} The records comprise 33 executed task-outcome contrasts, 49 comparisons with restricted endpoints, and 12 diagnostic summaries. Only 14 task contrasts report contrast-level uncertainty, and the response policy is documented as retained in nine, all rule replacements from four studies; none of the 12 necessity-decision records executes a task contrast. Source checks distinguish whole-plan correctness, correct skill loading, and choice accuracy conditional on presentation from completed workflows. In Outcome Monitors, a count-matched timing placebo detects no timing effect (+0.88 points, 95\% CI \ensuremath{-}4.39 to 6.14), so the source evidence attributes the detectable package gain to recovery content; detector selection at a fixed warning quota remains untested.

{\bf Conclusions:} The analysis separates evidence that can inform adoption from evidence that supports component development. Recovery and disruption, intervention timing, and downstream consumption organize three candidate mechanisms, and eight conditional reporting items specify the comparisons needed to distinguish the remaining explanations.
\end{abstract}
\maketitle
\section{Introduction}

A language-model agent repeatedly selects tools, routes steps to models, requests assistance, and decides whether to continue or intervene. Replacing one such decision procedure is a local engineering change with potentially nonlocal consequences. The decision shapes the execution trajectory: withholding a tool changes the available actions, a warning changes what the agent observes, and a restart changes which state and resources remain available. Evaluating the replacement therefore requires connecting a decision to what the agent subsequently does.

Outcome Monitors provides a concrete example. Its main comparison adds a monitor and a recovery receipt to a fault-affected workflow and reports improved completion. The receipt supplies information that the agent can use after a warning. A separate always-warn comparison varies the recovery-tool list while retaining the warning envelope, and a preregistered placebo shifts warnings to other positions at a matched count. The first comparison evaluates the package; the second shows that recovery content matters; the third finds no detectable contribution from warning timing. Detector selection at a fixed warning quota is the one contrast the source leaves untested \citep{ref1}. This distinction matters when deciding whether to deploy the package, improve its detector, or change the actions available after detection.

We distinguish \textbf{Question A: does the specified replacement improve the task-level outcome under the tested conditions?} from \textbf{Question B: do changes in local decision quality explain that outcome?} A controlled comparison with executed outcomes can estimate A even if no local-quality metric is reported. B requires a relevant local measurement under comparable conditions and a design that distinguishes the decision rule from accompanying changes in information, budget, and response policy. Intervention frequency can itself be part of the rule's effect; holding frequency fixed answers the narrower question of which interventions to select at a given quota. A favorable local score cannot substitute for task measurement, and aligned local and task improvements do not by themselves establish attribution.

The research question is what existing replacement comparisons establish about task benefit and local-quality attribution, and how those conclusions should guide agent design. We address it through three contributions.

\begin{enumerate}
\item A five-step interpretation procedure separates evidence for adopting a replacement from evidence for improving a decision component. Its application to every retained record produces a source-linked assessment of the intervention, endpoint, supported conclusion, design boundary, information gap, and next comparison.
\item The record-level synthesis shows where the evidence for adoption and attribution actually lies: which decision categories supply executed task contrasts, how many of those contrasts resolve their uncertainty, and how rarely the response policy is held fixed. Eight worked cases connect these findings to design choices, including a beneficial monitoring package, offline localization, and a termination safeguard. A contrast decomposition defines detector value and explains why effects from different populations and warning policies cannot be combined into a detector attribution.
\item Cross-case synthesis links three candidate mechanisms to eight conditional reporting items and to comparisons that could distinguish the remaining explanations. The resulting guidance specifies what to retain, vary, and measure when assessing a rule, its response policy, or their combination.
\end{enumerate}

The 348-study map establishes the coverage of this review; the interpretive synthesis uses 94 comparison records and purposively selected cases. Section 2.4 states the inferential scope of these records.

\subsection{Relationship to adjacent reviews}

Recent reviews cover the broader evaluation landscape: Nageshwaran and colleagues examine methods, benchmarks, and interaction validity; Kehkashan and colleagues connect evaluation to deployment, including process quality, recovery, cost, and long-horizon attribution \citep{ref2,ref3}. Varangot-Reille and colleagues organize routing by objective, timing, and implementation, and discuss controls that separate candidate-pool quality from router contributions \citep{ref4}. These reviews establish the relevance of interaction, recovery, and controlled comparison.

Our unit of synthesis is the replacement--comparator contrast within a study. This makes it possible to distinguish two results from the same paper that answer different design questions. The Outcome Monitors analysis, for example, traces the populations and policies behind its package comparison and recovery-content control before asking what either establishes about detector quality. The analysis of First-chunk selection and Evidence-Carrying Termination (ECT) distinguishes the task actually measured from the task for which a component is intended. These are assessments of reported evidence that complement the broader taxonomies and evaluation principles above.

Plaat and colleagues provide the broader agent-capability context, while Mandi and colleagues establish the prediction--decision distinction in decision-focused learning \citep{ref5,ref6}. Frauen and colleagues frame changes to routing, retrieval, and agentic workflows as causal interventions and discuss identification and credit assignment across components \citep{ref7}. We use these established distinctions to interpret the measurements and controls in the reviewed studies. The additional result is a comparison-specific account of which explanations remain possible and which next contrast could distinguish them. In an executing agent, the decision also changes subsequent observations, retained state, and available recovery actions; Sections 3.5 and 4.1 follow these consequences through the source evidence. The decomposition below organizes these empirical judgments without claiming a new causal identification theorem.

\section{Methods}

\subsection{Scope and eligibility}

We conducted a critical scoping review of separable decision components intended for multi-step language-model agents that use tools or act in an environment. Eligible studies replace an identifiable decision component and report a local metric or downstream endpoint. Offline component studies are eligible when designed for this setting; their evidence is distinguished from closed-loop execution. We coded the selection of an entire agent configuration before a task separately from decisions made during execution.

General foundation-model training, undifferentiated prompting changes, pure text-only reasoning ensembles, memory organization, and studies solely concerned with attack prevention are outside scope. An execution gate evaluated through task completion can remain relevant even if it also reports a safety endpoint. Ambiguous cases are retained with explicit qualifications and sensitivity analyses.

\subsection{Information sources and selection}

We searched arXiv, OpenAlex, Semantic Scholar citation links, ACL Anthology, and OpenReview on 24 September 2026, and Scopus through institutional access on 25 September. Two arXiv search rounds used agent, language-model, and control-action terms, with an outcome block in the first round, and covered computer-science submissions from 1 September 2023 to 24 September 2026. The Scopus search used title/abstract fields, restricted the subject area to computer science, and covered publication years later than 2022; the export did not allow verification of a month-level cutoff.

Deduplication, title screening, keyword rechecks, abstract screening, and eligibility checks yielded 348 studies: 134 from the first arXiv round, 125 from the second, 24 from supplementary searches, and 65 from Scopus. Figure 1 summarizes selection. Appendix A.1 documents counts for each search source, screening checks, coverage limits, and available logs.

\begin{figure}[!htbp]
\centering
\includegraphics[width=\textwidth]{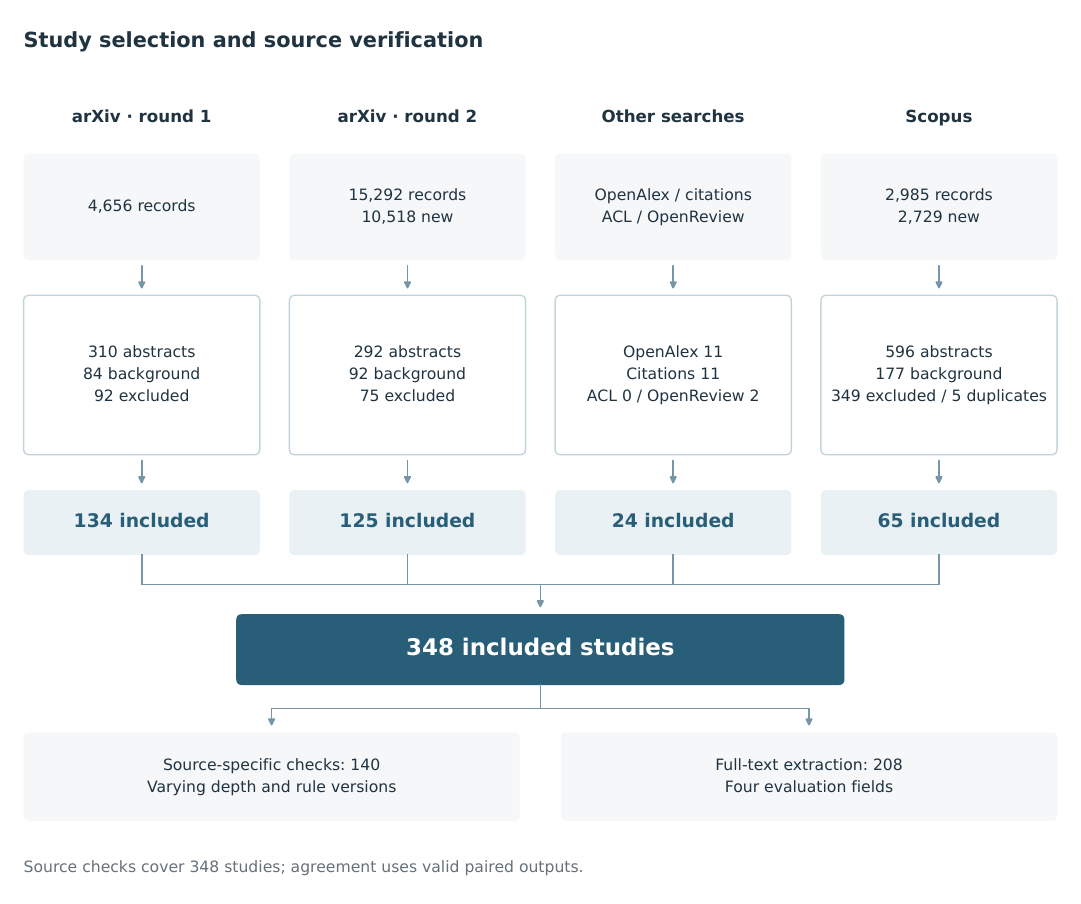}
\caption{Search-arm selection and verification coverage. Source-specific checks covered 140 studies and model-assisted paired extraction the other 208; the two differ in depth and coding rules and together do not constitute a sample independently double-coded under uniform conditions. Stage counts for each arm are given in Appendix A.1, and record-level screening decisions in the supplement.}
\Description{Flow diagram with four search arms side by side: arXiv round 1 (4,656 records, 310 abstracts, 134 included), arXiv round 2 (15,292 records, 10,518 new, 292 abstracts, 125 included), other searches (OpenAlex 11, citations 11, ACL 0, OpenReview 2; 24 included) and Scopus (2,985 records, 2,729 new, 596 abstracts, 65 included). The arms merge into 348 included studies, which split into 140 with source-specific checks and 208 with full-text extraction of four evaluation fields.}
\end{figure}

\subsection{Study-level coding and verification}

\textbf{Initial coding.} Each of the 348 included studies has a record identifying its primary decision point, controller type, and four evaluation fields: local decision quality, task endpoint, evaluation level, and multiple-backbone testing. Initial codes were assigned from abstracts and other available study information; source-based eligibility checks also supplied codes for included records. Appendix A.2 defines the ten decision categories and field rules.

A local metric measures the replaced decision's quality. Task endpoints distinguish measured success or completion (Y), a proxy (P), and no reported endpoint (N). Evaluation levels distinguish static/offline prediction (E0), replay, simulation, or offline prefix evaluation (E1), and online closed-loop comparison (E2). F denotes insufficient information. These study-level codes describe evidence coverage; they do not establish that local and task measurements come from the same comparison.

\textbf{Source verification.} Source-specific checks covered 140 studies with varying depth, including full texts, excerpts, and retained notes. The remaining 208 studies underwent model-assisted full-text extraction of the four evaluation fields. Extraction followed the evidence supporting the main claim rather than the highest evaluation level mentioned in the paper.

\textbf{Adjudication and correction.} Disputed values were adjudicated using recorded coding rules and source evidence; ambiguous decision-point and controller assignments received targeted checks. Current coding tables incorporate these decisions; original outputs and corrections linked to specific sources remain traceable. Agreement was calculated from valid pairs of pre-adjudication outputs, separately from the final adjudicated codes. A retrospective rule-compatibility check addressed earlier records; Appendix A.3 reports its scope alongside the agreement sample.

\subsection{Comparison-level extraction and synthesis}

A separate comparison-level dataset supports the inferential analysis. It contains 94 records from 42 studies: 92 from 40 included studies and two supplementary records. A study can contribute several replacement contrasts or associated diagnostics, including cross-setting summaries and component-only evaluations. Each record retains the replacement, comparator, backbone, task subset, measurements, denominators, repetitions, uncertainty estimates, tests, and source locations. Local-quality measurements are distinguished from resource use, intervention frequency, and downstream outcomes; study-level co-reporting alone does not establish a matched comparison.

The records come from four sources. Fifty-eight records from 15 studies are retained extracts from an earlier comparison re-audit; 30 records from 25 studies were added to check specific claims in drafts of this manuscript; two come from publisher full texts supplied later; and four are partner records added in the present revision because a retained source already reported them. Records added to check claims carry a risk of circularity, since the manuscript's arguments influenced which comparisons were extracted. By decision category, the records comprise 37 selection, 13 termination, 13 monitoring, 12 necessity, six gating, four configuration, four routing, and three deterministic-replacement records, plus the two supplementary records; no assistance or budgeting study contributed a record. The largest single-study contribution is seven records.

Eight cases were purposively selected for detailed synthesis across three candidate mechanisms, positive net-benefit evidence, attribution and deployment questions, and proxy or diagnostic endpoints. Seven are included studies; the budget study is supplementary. We applied the procedure in Section 2.6 to every record and reconciled each record whose arms, endpoint, or paired values were inconsistent against its source text. The annotations preserve the original values, record every corrected field alongside its historical value, and distinguish a documented design boundary from an extraction gap. Appendix A.4 documents the decision rules, record semantics, and supplementary-study membership. Search, coding, and comparison records accompany the review; reporting was checked against PRISMA-ScR~\citep{ref8}.

\textbf{Scope of inference.} The comparison records are a purposive collection, so their counts describe what the retained evidence permits, not how often the 348 studies or the wider literature establish task benefit or attribution. We do not pool effects or estimate a corpus-wide rate of local--task decoupling. The worked cases demonstrate how to apply the procedure; they do not evaluate whether it improves readers' accuracy, agreement, or speed. We conduct no new model experiments. The decomposition in Section 2.5 organizes empirical judgments and does not estimate unobserved conditions.

\subsection{Separating the decision rule from its response policy}

For the comparison synthesis, let $d$ denote the rule that selects a local decision and $r$ the policy that determines its consequences, including the information supplied and the recovery actions made available. Let $\mu(d,r)$ be the mean task outcome obtained by executing that combination under a specified task distribution, executor, endpoint, and budget allowance. Subscripts 0 and 1 index the comparator and replacement settings. A package comparison concerns

\[
\Delta_{\mathrm{package}}=\mu(d_1,r_1)-\mu(d_0,r_0).
\]

Adding and subtracting the same intermediate condition gives

\[
\Delta_{\mathrm{package}}=\underbrace{\mu(d_1,r_1)-\mu(d_0,r_1)}_{\text{rule effect under }r_1}+\underbrace{\mu(d_0,r_1)-\mu(d_0,r_0)}_{\text{response effect under }d_0}.
\]

This identity does not require additive effects. It requires the same intermediate condition and common evaluation setting for the terms to describe one package contrast. Subtracting a response effect measured under another rule or task distribution does not recover the first term without additional assumptions. A component's effect may also depend on its partner: comparing the rule contrast under both response policies requires the fourth condition, $\mu(d_1,r_0)$. When the components interact, using this alternative intermediate condition changes how the gain is divided between the two terms. Even a controlled rule effect is not automatically an effect of measured decision quality; that attribution still needs a relevant quality measure and a design addressing other differences between the rules.

The decomposition is uninformative when the comparator never triggers. If $d_0$ is a no-warning rule, the response policy has no occasion to act under it, so $\mu(d_0,r_1)=\mu(d_0,r_0)$, the response term is zero, and the whole package effect appears as a rule effect. Detector value is therefore defined only against an alternative trigger rule, such as always-warn, a random or timing placebo, or a simpler detector, under a common response policy and, when the claim concerns selection, a common warning quota $q$:

\[
\Delta_{\mathrm{rule}}(d_{\mathrm{det}}\ \mathrm{vs}\ d_{\mathrm{alt}}\mid r,q)=\mu(d_{\mathrm{det}},r;q)-\mu(d_{\mathrm{alt}},r;q).
\]

The distinction also determines which factors a control should retain. Holding the response policy fixed preserves what a warning enables, while allowing a different decision rule to warn on different trajectories. Matching the number of warnings instead estimates a rule comparison at that quota. Neither design makes treatment-dependent triggered groups comparable by itself. We use these distinctions to interpret existing contrasts and identify missing comparisons; we do not estimate unobserved conditions or component interactions from the study-level codes.

\subsection{A procedure for interpreting a replacement comparison}

We apply five steps to connect a reported result to an engineering decision. The output is a supported conclusion, its scope, and the comparison or information needed to answer the remaining question.

\begin{enumerate}
\item \textbf{Specify the decision.} State whether the objective is task improvement (A), attribution to local quality (B), performance preservation, lower cost, or component development. Define the task population and outcome before interpreting a favorable score.
\item \textbf{Reconstruct the intervention.} Identify the replacement and comparator, separating the decision rule from its response policy. Record which executor, information, population, and budget allowance are common. A rule can change its outputs and trigger frequency under a fixed response policy; those changes are not automatically confounds. Check whether the source already reports a partner, placebo, or quota-matched control before proposing one.
\item \textbf{Locate the measurement.} Read the endpoint definition, unit, denominator, and evaluation protocol. Distinguish a completed task from selection, loading, plan correctness, or saved replay steps. Check whether the changed policy was executed and whether local and task measurements concern the same conditions. An execution label alone does not establish an endpoint's meaning.
\item \textbf{State what follows.} Retain positive and adverse results at the scope actually tested. Separate an observed point difference, an uncertainty-supported gain, a preservation result under a stated margin, and an inconclusive comparison. Apply the shared-condition requirement in Section 2.5 before combining component effects. Two improving metrics do not identify a quality-mediated effect.
\item \textbf{Identify what would resolve the question.} Name the missing partner control, executed endpoint, population alignment, or uncertainty estimate. If the extract cannot identify the arms or endpoint consistently, request reconciliation of that record before proposing a new experiment. Missing extraction fields do not establish that the source omitted a comparison.
\end{enumerate}

For the record-level assessment, each record receives one reading scope: an executed task-outcome contrast; a comparison restricted to a component, proxy, offline/replayed or simulated endpoint; or a diagnostic summary without a common replacement contrast. A record whose arms or endpoint cannot be reconciled from the extract is checked against the source before a scope is assigned. These scopes determine the question that can be assessed, not whether the result is favorable or the study is well designed. A live agent acting in a deterministic tool environment can supply task-outcome evidence; a simulation that substitutes probabilities for agent behavior cannot establish that same execution claim. Each annotation also records the apparent rule, response, or package change and the evidence for retained response conditions. For task contrasts it further records whether the source reports an interval or test for the contrast itself, only arm-level variation, or neither, and the direction of the result at that level of support. The procedure is analytical guidance illustrated by the cases, not a validated decision aid.

\subsection{Use of language models}

Language models assisted the initial abstract screen of Scopus records, whose 305 proposed inclusions the authors reviewed (Appendix A.1), the four-field extraction for 208 studies through two coding streams, adjudication of disputed fields, the retrospective rule check, and the record-level annotations of Section 2.6, including their reconciliation against sources. The rule check used gpt-5.6-terra to review sources after an initial screen with deepseek-v4-flash; the source reconciliation of the comparison records in October 2026 used Claude Opus 5.5. Models also assisted literature retrieval, claim verification against sources, language editing, and generation of the illustrative Figure 3. The authors defined the scope, coding rules, and procedure, selected the worked cases, and are responsible for every reported judgment. The record-level annotations were not double-coded by humans.

\section{Results}

\subsection{Corpus composition and evidence levels}

The 348 studies were coded as 84 selection/shortlisting decisions, 56 routing decisions, 49 monitoring interventions, 41 gates, 32 termination/restart decisions, 27 task configurations, 24 necessity decisions, 14 assistance decisions, 12 budgeting decisions, and nine deterministic replacements. Controller types were coded as rules/algorithms in 109 studies, classifiers in 67, language-model judges in 65, retrievers in 49, reinforcement-learning/bandit policies in 23, probes in 19, and smaller models in 16.

The categories also differ in what is delegated. Agentic Metacognition uses failure triggers to initiate a human handoff (ASK), whereas To Call or Not to Call studies whether a tool invocation is needed (NEC) \citep{ref9,ref10}. Ares selects reasoning effort at each step (BUD); TraceCompiler compiles recurring traces into mostly deterministic workflows with residual language-model decisions (REP); and Strategy Auctions selects among agents using short task-level plans (CFG) \citep{ref11,ref12,ref13}. These examples clarify which decision is changed; they do not establish that the evaluations provide equally strong evidence of downstream benefit.

The complete category index is in Appendix B; stable study identifiers link it to the supplementary included-study catalogue.

The current coding assigned E0 to 54 studies, E1 to 61, and E2 to 229; 4 remained insufficiently specified. These levels describe evaluation modes, rather than establish the validity of every causal interpretation drawn from them. In particular, a system can execute online while changing several components at once.

Figure 2 shows the distribution of current evaluation-level codes across decision points.

\begin{figure}[!htbp]
\centering
\includegraphics[width=\textwidth]{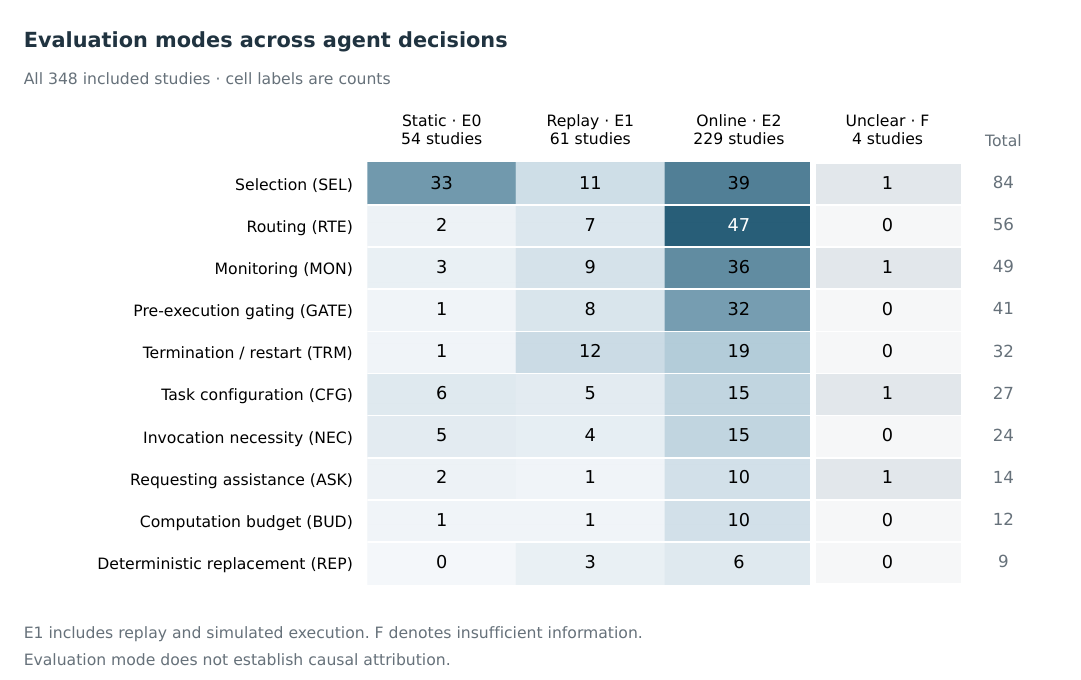}
\caption{Current study-level coding across decision points and evaluation levels (N = 348). F denotes insufficient information and is shown separately. All studies are retained; cell counts describe the corpus, not the adequacy of causal evidence.}
\Description{Heat map of 348 studies with ten decision points as rows and four evaluation levels as columns (static E0, replay E1, online E2, unclear F). Online E2 dominates most rows; selection has the most static studies (33); routing has the most online studies (47). Row totals range from 84 for selection to 9 for deterministic replacement.}
\end{figure}

\subsection{Co-reporting does not establish a matched relationship}

\begin{table}[!htbp]
\caption{Co-reporting of local decision metrics and task endpoints (N = 348).}
\centering
\small
\begin{tabular}{lrrrr}
\toprule
Local metric & Task Y & Task P & Task N & Task F \\
\midrule
Reported Y & 142 & 49 & 31 & 0 \\
Not reported N & 113 & 12 & 0 & 0 \\
Insufficient F & 0 & 0 & 1 & 0 \\
\bottomrule
\end{tabular}
\end{table}

Local metrics were reported by 222/348 studies (63.8\%); 125 were coded N and 1 F. Most studies in the current corpus therefore reported a local metric. The relevant evidence gap concerns comparability and attribution. The 142 studies also reporting measured endpoints and the 49 also reporting proxies form a pool of 191 studies for comparison-level analysis. Reporting two metrics somewhere in a paper does not show that the same replacement produced both measurements.

A local metric was reported without a task endpoint in 31 studies, including 16 SEL studies. Examples of retrieval-only evidence include Field-Aware Skill Retrieval, R3-Skill, and DSR \citep{ref14,ref15,ref16}. Their component measurements can be informative, but compatibility with a benchmark does not show that the agent task was executed.

Multiple-backbone testing was coded Y/N/F in 178/159/11 studies; 135 E2 studies included tests on multiple backbones. These counts do not measure transfer of a fixed controller. After excluding 11 studies identified as falling near the scope boundary, the sensitivity analysis still found that a majority reported a local metric. Appendix A.5 reports the subset counts and their interpretation.

\subsection{Coding checks}

Across the four evaluation fields, agreement between valid pairs of pre-adjudication outputs was 82.7--91.4\%. Appendix A.3 reports field-specific denominators, batch-level agreement, and changes from initial codes. These checks describe coding consistency under the recorded rule versions; the corpus statistics use the adjudicated and corrected tables.

\subsection{Applying the procedure to all retained records}

The assessment assigns the 94 records to three reading scopes (Table 2). Thirty-three records from 20 studies describe executed task-outcome contrasts, 49 from 24 studies compare restricted endpoints or evaluation protocols, and 12 from seven studies provide diagnostics without a common replacement contrast. Nine historical records could not be interpreted from the extract alone; reading each against its source resolved all nine, three as task contrasts, five as restricted comparisons, and one as a diagnostic.

\begin{table}[!htbp]
\caption{Reading scopes for the 94 comparison records. Counts describe this purposive collection, not study quality or prevalence in the corpus.}
\centering
\footnotesize\setlength{\tabcolsep}{3pt}
\begin{tabular}{@{}>{\raggedright\arraybackslash}p{0.22\textwidth}>{\raggedright\arraybackslash}p{0.07\textwidth}>{\raggedright\arraybackslash}p{0.32\textwidth}>{\raggedright\arraybackslash}p{0.32\textwidth}@{}}
\toprule
Reading scope & Records & Supported use & Unresolved question \\
\midrule
Executed task-outcome contrast & 33 & Assess the specified rule, response, or package on its measured task outcome. & Whether the difference is resolved by uncertainty, transfers to another setting, or is attributable to local quality. \\
\addlinespace[4pt]
Restricted endpoint or evaluation & 49 & Retain component, proxy, replay, simulation, or other limited comparisons at their stated scope. & Whether the result persists when the changed policy executes the intended complete task. \\
\addlinespace[4pt]
Diagnostic summary & 12 & Use cross-setting ranges, stage profiles, measurement changes, and single-method diagnostics to locate a question. & What a common-population replacement contrast would establish. \\
\addlinespace[4pt]
\bottomrule
\end{tabular}
\end{table}

Table 3 shows where these scopes fall. Task contrasts concentrate in selection (11 records from six studies) and monitoring (nine from five). The 12 necessity records, from five studies, include no executed task contrast, and termination contributes one among 13. Routing has four records and assistance and budgeting none, so the collection says little about whether those decisions improve task outcomes.

\begin{table}[!htbp]
\caption{Comparison records by decision category and reading scope. Records added to check manuscript claims are included; counts are not prevalence estimates.}
\centering
\small
\begin{tabular}{lrrrrrr}
\toprule
Decision category & Records & Studies & Task & Restricted & Diagnostic & Studies with task \\
\midrule
Selection and shortlisting & 37 & 13 & 11 & 25 & 1 & 6 \\
Termination and restart & 13 & 4 & 1 & 7 & 5 & 1 \\
Monitoring and intervention & 13 & 7 & 9 & 3 & 1 & 5 \\
Necessity decisions & 12 & 5 & 0 & 10 & 2 & 0 \\
Gating & 6 & 3 & 4 & 0 & 2 & 3 \\
Task configuration & 4 & 3 & 2 & 2 & 0 & 1 \\
Routing & 4 & 4 & 2 & 2 & 0 & 2 \\
Deterministic replacement & 3 & 1 & 3 & 0 & 0 & 1 \\
Supplementary studies & 2 & 2 & 1 & 0 & 1 & 1 \\
\midrule
All records & 94 & 42 & 33 & 49 & 12 & 20 \\
\bottomrule
\end{tabular}
\end{table}

Within the 33 task contrasts, 15 replace a rule, 15 change a larger package, and three vary a response policy. Fourteen, from ten studies, report an interval or test for the contrast itself: nine favour the replacement, one supports completion non-inferiority, three are inconclusive, and one favours the comparator. The other 19 rest on point estimates, five of them with arm-level variation only; 13 favour the replacement, four the comparator, one is a tie, and one reverses between validation and test splits. Most favourable task evidence in the collection is therefore a point-estimate gain.

Attribution needs more than an executed contrast. The response policy is documented as retained in nine task contrasts, all rule replacements from four studies: State-Path menus, HYSET, SchemaRouter, and the Outcome Monitors placebo and always-warn controls \citep{ref17,ref18,ref19,ref1}. Every package contrast changes or only partly specifies the response conditions, so those 15 records can support adopting the package (A) but not attributing its gain to a decision component (B). Only three of the nine documented-response contrasts also report contrast-level uncertainty. Local and task measurements move apart in two of the nine: COLT has higher Recall@5 than ToolLLaMA-Ret but a slightly lower judged pass rate, and SchemaRouter's minimal-field work point has higher field F1 (0.75 versus 0.59) but lower answer accuracy (0.56 versus 0.71) with identical tool selection.

Endpoint checks reveal why a comparison can remain restricted even when its historical extraction mentions execution or success. Dynamic Tool Dependency Retrieval scores entire plans for correct function names and parameters, while its planning protocol does not interleave tool execution. Gavel's Skill-Use result measures whether the correct skill is loaded during execution. E-valuator calculates saved tokens against the continuation of collected trajectories. These results respectively inform plan construction, skill activation, and retrospective stopping tradeoffs; they do not supply the same evidence as measured completion after executing the changed workflow (\citealp[\S{}\S{}3, 6]{ref20}; \citealp[Table 1]{ref21}; \citealp[\S{}4.3]{ref22}). Choice accuracy can mislead in the same way. BoR's adaptive shortlist raises the rate at which the gold tool is chosen when it is presented (93.1\% versus 87.1\%) but lowers end-to-end accuracy against a fixed five-tool list (71.7\% versus 73.3\%), because the two arms present the gold tool on different query subsets (\citealp[Table 1]{ref23}).

Across scopes, the annotations also preserve evidence that limits a favorable mechanism account: the budget comparison favors extra baseline steps in its reported average, and Failure Prediction includes an adverse executed intervention. Point reversals without a resolving interval remain point reversals. Section 3.5 develops the cases, and Section 4.1 applies the procedure to positive and boundary results in full.

\subsection{What the worked cases establish}

We examine each selected case through five questions: which decision or package was replaced, which comparison was executed, whether local and downstream measurements refer to comparable conditions, which explanations the controls rule out, and which remain. A denotes net benefit of the stated replacement and B attribution to local decision quality. Section 2.5 distinguishes changes to the decision rule from changes to its response policy. A claim about the full task requires a task endpoint measured under the changed execution policy; an offline comparison can still address a narrower question.

Table 4 applies this structure to eight cases. Each row shows how the evidence supports a particular inference, sometimes drawing on separately identified contrasts; it is not an independent effect estimate. The rows span positive net benefit, attribution limits, proxies, and action-value diagnostics. They are selected from the comparison collection described in \S{}2.4; the budget case is supplementary. M1--M3 refer to the three candidate mechanisms synthesized below; attribution and deployment questions are developed in Section 4.

\begingroup\footnotesize\setlength{\tabcolsep}{3pt}
\begin{longtable}{@{}>{\raggedright\arraybackslash}p{0.18\textwidth}>{\raggedright\arraybackslash}p{0.25\textwidth}>{\raggedright\arraybackslash}p{0.25\textwidth}>{\raggedright\arraybackslash}p{0.25\textwidth}@{}}
\caption{Selected comparison evidence by use: net benefit (A), attribution (B), and interpretation boundaries. Seven included studies and one supplementary study; purposive selection, not prevalence estimation.}\\
\toprule
Replaced decision and evidence use & Executed comparison and controls & Measurement comparability & Supported claim and remaining explanation \\
\midrule
\endfirsthead
\toprule
Replaced decision and evidence use & Executed comparison and controls & Measurement comparability & Supported claim and remaining explanation \\
\midrule
\endhead
\bottomrule
\endlastfoot
A and B boundary; M1. Failure prediction \citep{ref24} & Paired intervention and baseline outcomes on the same tasks; task-level resampling and repeated seeds. & Critic AUROC 0.936 on held-out examples; task success decomposed into recovered failures and disrupted successes. & A: net effects can be estimated within tested conditions. B: predictive discrimination alone does not identify the benefit of intervention. Classifier samples and executed tasks have different denominators. \\
\addlinespace[4pt]
A with policy control; M2. RestartSmart \citep{ref25} & 100 test instances and five seeds; no-intervention and cold-restart controls at the stated warning operating point. & A 25\% target false-positive rate is an operating specification, not a measured quality improvement. Task resolution: 66.6\% baseline, 71.8\% smart restart, 66.8\% cold restart. & A: the smart-restart gain has a +2.8 to +7.6-point trajectory-level bootstrap interval, and paired McNemar tests favour it over both controls. B: recovery policy matters; no within-study manipulation isolates reversibility. \\
\addlinespace[4pt]
Aligned measurements; M3. HYSET matched feedback \citep{ref18} & ToolGen versus HYSET with matched execution feedback, judge, training budget, and fixed executor. & Recall@5: 83.12\% to 84.75\%; judged task pass rate: 66.85\% to 69.69\%. & Both point estimates improve. B: architecture and retrieved context still differ; contrast-specific uncertainty is not verified. Significance reported in the main table cannot be applied to this contrast. \\
\addlinespace[4pt]
Proxy boundary; M3. First-chunk selection \citep{ref26} & FIFO versus prioritized chunks; 100 tasks with reachable gold files; paired tests within each model. & Gold-first placement: 50\% to 70--72\%; offline file localization changes by \ensuremath{-}2.8 to +2.2 points across five models, with accuracies near ceiling (81.8--94.0\%). & No significant localization difference was detected. This is neither equivalence nor A for closed-loop issue resolution. Interpretation is limited by the reachability criterion, the ceiling, and the proxy endpoint. \\
\addlinespace[4pt]
A versus content attribution. Outcome Monitors \citep{ref1} & Paired monitor-plus-receipt comparison; on 57 further workflows and two tiers, an always-warn control varies recovery-tool lists and a preregistered placebo shifts warning timing at a matched count. & No paired local-quality contrast. Main workflow completion: 10.9\% to 28.1\%; full versus stripped receipts: 30/114 to 43/114; detector timing versus placebo: +0.88 points. & A: the complete package improves completion in the tested fault setting. B: the detectable gain is carried by recovery content; timing shows no detectable effect; selection at a fixed quota is untested. \\
\addlinespace[4pt]
Budget control. Online skill/memory study, supplementary \citep{ref27} & Vanilla-IB versus ASI with budget-aware horizon allocation; Gemini 3 Flash; three WebArena domains and three runs (v1). & No local-quality comparison. Tokens/task: 71.9K versus 107.1K (resource use); success: 50.74\% versus 47.86\%. & A: point estimates motivate comparing augmentation with extra baseline steps. B: neither equal realized costs nor a quality mechanism is isolated; the reported average is shown without an interval. \\
\addlinespace[4pt]
Action-value and label boundary; M1. Calibration Is Not Control \citep{ref28} & Same-prefix counterfactual branches; scalar trigger versus action-conditioned controller on ALFWorld. & Control regret: 0.506 versus 0.110. This action-value measure also serves as the reported utility endpoint, rather than providing a second, independent task-success metric. & B: failure-risk prediction and intervention value are different targets. Offline branching supports this distinction; A for a complete online controller is not established by this contrast. \\
\addlinespace[4pt]
Bounded positive evidence; M1. Evidence-Carrying Termination \citep{ref29} & Same planner configuration, prompt, tools, and nominal checkpoint; independent planner calls per arm. Three seeds, task-cluster intervals, and a prespecified \ensuremath{-}10-point non-inferiority margin. & Premature stopping: faithful controller 40/66, ECT 0/66; supported completion: ECT 97/132, faithful controller 92/132. & Completion non-inferiority is supported; completion superiority (A on this endpoint) is not established. Reduced premature unsupported termination is a separate positive endpoint. B: the measures have distinct denominators, and their alignment does not isolate the contribution of decision quality. Synthetic tools and the substitute model limit transfer. \\
\addlinespace[4pt]
\end{longtable}
\noindent\begin{minipage}{\linewidth}\footnotesize\raggedright
Abbreviations: AUROC, area under the receiver operating characteristic curve; FIFO, first-in, first-out; ASI, Agent Skill Induction; Vanilla-IB, vanilla baseline with an increased budget.
\end{minipage}\par\medskip
\endgroup

This structure explains how a favorable system comparison can support A while leaving B unresolved, and how an offline or proxy comparison can inform a mechanism without establishing A. Positive findings retain their value within these distinctions. The mechanisms are interpretive categories that can overlap within a case; their frequency in this selected table does not estimate their prevalence in the literature.

\textbf{Interference and recovery.} An intervention can rescue a trajectory that would otherwise fail or disrupt one that would otherwise succeed. When paired potential outcomes can be evaluated, let p be baseline failure probability, r recovery conditional on baseline failure, and d disruption conditional on baseline success. The net change is \ensuremath{\Delta}Success = p r \ensuremath{-} (1 \ensuremath{-} p)d. This decomposition explains why the same warning quality can have different value when the baseline or recovery policy changes; it does not assume that r and d transfer between settings. The paired analysis in Failure Prediction captures both outcomes, whereas Calibration Is Not Control uses same-prefix branches to distinguish failure risk from the value of a particular intervention \citep{ref24,ref28}.

These cases distinguish two evaluation targets: detecting a bad state and selecting a beneficial response. Their endpoints delimit this interpretation: Calibration's control regret measures action value rather than independent online completion; Verifier Tax compares proposal interception with episode-level safe success without a paired local-quality baseline; and SU veto's opposing step-error and task-resolution point estimates are reported without estimates of repeated-run uncertainty \citep{ref30,ref31}. These observations motivate measuring recovery and disruption under a common task distribution. They do not establish a general harmful effect of accurate monitoring. To test that explanation, a comparison would need to vary the decision rule while retaining the intervention content and measuring both rescued failures and disrupted successes.

A complementary simulation result separates containment from completion. In Containing the Cascade, removing checkpoint rollback leaves containment at 99.5\% but reduces simulated task success from 83.3\% to 2.6\% (Table V). Stopping propagation can therefore preserve one objective while preventing progress toward another; recovery is part of the intervention, rather than a consequence guaranteed by detection \citep{ref32}.

\textbf{Timing and reversibility.} A warning can guide action only through the options available when it is issued. Last Step Matters reports stronger confidence discrimination at the end of a trajectory than midway through it, with paths also changing; RestartSmart compares an edit-preserving restart with an optional diff overlay against both continued execution and a cold restart \citep{ref33,ref25}. Read together, these cases separate the ability to predict an eventual outcome from the ability to improve it before recoverable state is lost. RestartSmart's control comparison shows that recovery policy matters within its coding setup. The timing evidence explains why a late, accurate signal need not offer the same opportunities for intervention as an early signal.

This interpretation applies when the intervention can preserve, restore, or discard task state. Neither study independently manipulates reversibility while holding model, task, and intervention policy fixed; Last Step Matters' Pareto comparison is qualitative. Differences between the studies therefore suggest reversibility as a candidate moderator without identifying its effect. A comparison designed to test this interpretation would vary the intervention checkpoint and state-retention policy within one environment, recording remaining cost, recovery, disruption, and completion.

\textbf{Mismatch between a local target and downstream use.} Ranking an option correctly, making it available to the agent, and using it successfully are distinct events. HYSET's retrieval and judged-pass comparisons, First-chunk's placement and localization comparison, and SkillApt's distinction between retrieving and loading a skill all expose this boundary \citep{ref18,ref26,ref34}. Together, these cases show why a local target must be interpreted in relation to the component that uses its output: a better shortlist can be ignored, consume context, or alter what the executor sees. These are candidate pathways, not effects isolated by all three studies.

HYSET's matched-feedback comparison reduces differences in supervision, but architecture and retrieved context still change; improvements in both point estimates do not isolate the contribution of retrieval quality. HYSET's own ablation shows the same separation inside one system: removing execution feedback lowers judged pass rate by 6.5\% relative but its set-level retrieval target, COMP@5, by only 0.7\% (\citealp[Table 2]{ref18}). Table 4 retains the historically extracted Recall@5; on COMP@5 the matched-feedback contrast is 70.94\% versus 77.55\%. First-chunk's paired tests concern an offline proxy and do not establish equivalence. SkillApt reports matching accuracy with reduced loading for activation on one benchmark, without an equivalence design. To distinguish these interpretations, a comparison would need to vary selection or activation while holding the executor, feedback, and budget fixed, and record both what the agent actually uses and the relevant task endpoint. Such a design can distinguish an unused local improvement from a useful replacement with an unresolved mechanism.

Two-Stage Tool Retrieval and Invocation offers a further distinction between efficiency and completion: in its GPT-4o airline evaluation, average tokens fall from 3364.43 to 2923.46 (13.1\%), while Pass1 changes from 0.420 to 0.425. The result supports an efficiency benefit with close success point estimates; it does not establish that retrieval-quality gains produce a corresponding increase in completion \citep{ref35}.

Figure 3 uses an illustrative invoice-email task to show the three pathways: rescuing or disrupting a trajectory, intervening while corrective actions remain available, and connecting selection to actual downstream use.

\begin{figure}[!htbp]
\centering
\includegraphics[width=\textwidth]{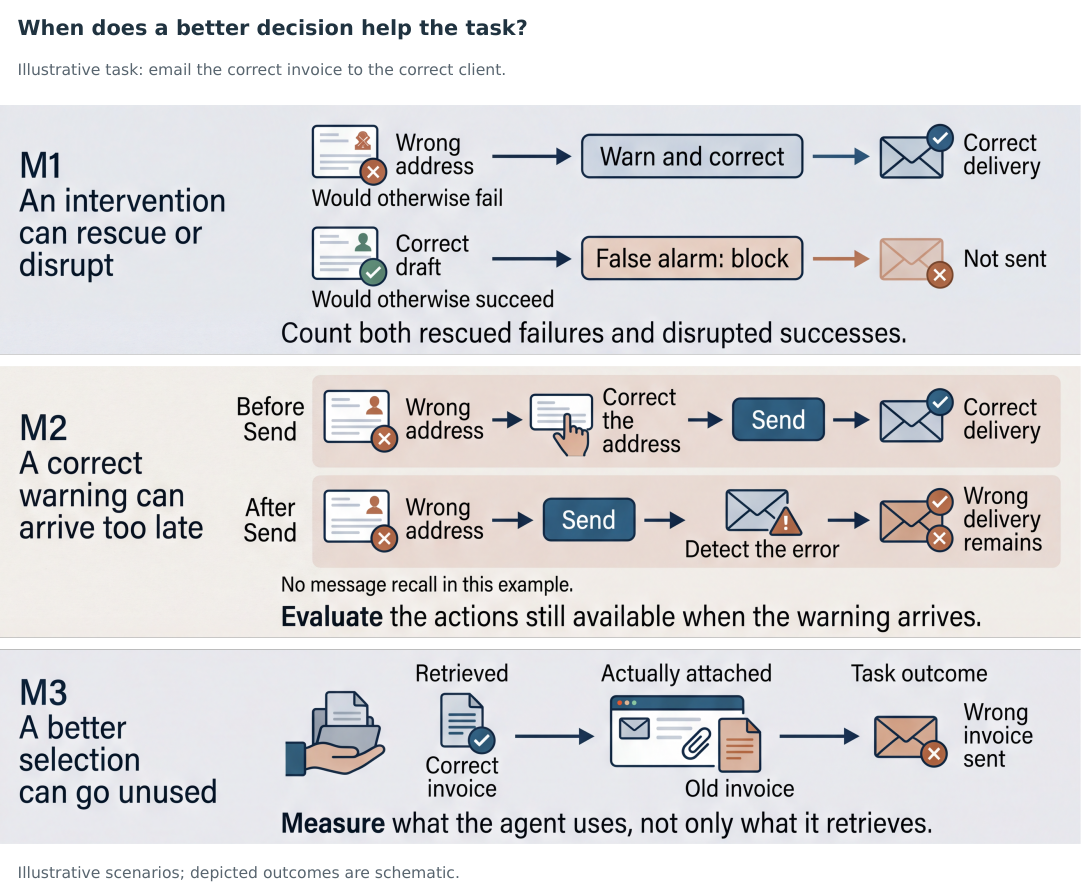}
\caption{Three candidate mechanisms illustrated through an invoice-email task. M1 contrasts correcting a wrong address with blocking a correct draft. M2 contrasts a warning before sending, when the address can still be corrected, with one after sending; message recall is unavailable in this example. M3 shows a correctly retrieved invoice going unused when the agent attaches an old invoice.}
\Description{Three illustrative panels using an invoice-email task. M1: a warning can rescue a would-be failure or block a would-be success, so both must be counted. M2: a correct warning before sending allows correction, while the same warning after sending cannot undo delivery. M3: a retriever finds the correct invoice but the agent attaches an old one, so retrieval quality does not reach the outcome.}
\end{figure}

\section{Discussion}

\subsection{From net benefit to attribution}

The first application asks whether to adopt \textbf{Outcome Monitors} and whether its detector explains the gain. Reconstructing the interventions identifies a monitor-plus-receipt package on 80 workflows across four models and, on 57 further workflows across two model tiers, a full-versus-stripped receipt control under always-warn triggering and a preregistered four-arm study of warning timing. The endpoint in each comparison is workflow completion: 35/320 to 90/320 in the package comparison, a reported gain of 17.2 percentage points with a task-cluster bootstrap interval of [11.25, 23.44]; and 30/114 to 43/114 in the receipt comparison, a reported gain of 11.4 points with an interval of [2.63, 20.18] (\citealp[Tables 2 and 18]{ref1}). The supported conclusions are package benefit within the tested fault setting and a contribution from recovery content under its own warning policy.

These two gains cannot be subtracted to estimate a detector contribution. In the notation of Section 2.5, the main comparator never warns, so the package effect appears entirely as a rule effect and says nothing about detector value on its own. The recovery-content contrast changes $r$ under an always-warn rule, whereas the main package uses a detector-triggered rule and a different evaluation population. The shared intermediate condition needed for the decomposition is absent, so the recovery-content interval cannot be read as a fraction of the main package's gain.

The source also reports the contrasts closest to detector value. In the four-arm study, every warned arm delivers a byte-identical generic caution inside the same envelope, including the recovery-tool list, and the arms differ only in when the warning attaches. Detector-timed warnings complete 33 of 114 workflow cells and a count-matched placebo, which delivers each suppressed warning at the next detector-consistent position, completes 32 (+0.88 points, 95\% CI \ensuremath{-}4.39 to 6.14; the source's minimum detectable effect is about 18 points; Appendix L, Table 17). Delivering the same envelope at every step completes 46, but this arm exposes recovery tools at about 600 events per tier, against 80--91 for the triggered arm, and the detector is silent in 9 of 57 workflows; the source attributes part of that margin to this exposure and to recall coverage. The learned-versus-schema-only ablation agrees: completion is 25/80 versus 21/80 (+5.0 points, p = .42), with different warning counts (111 versus 166) and receipt contents. Together with the receipt contrast, this evidence attributes the detectable gain to recovery content rather than to warning timing, while a timing contribution below about 18 points remains possible.

The placebo keeps the detector's choice of workflows, so it does not test selection. The comparison that remains is the detector against random or simpler selection at the detector's warning quota, with the same envelope: $\Delta_{\mathrm{rule}}(d_{\mathrm{det}}\ \mathrm{vs}\ d_{\mathrm{alt}}\mid r,q)$ in Section 2.5. The current evidence supports A for the package; it neither establishes nor rules out a contribution of detector selection to completion.

The same attribution question applies when a replacement changes computation or assignment as well as the rule. The supplementary budget study compares online skill or memory augmentation with extra baseline steps; its cited v1 evaluation covers Shopping, Reddit, and Admin. Content-insensitive routing controls ask whether assignment needs to depend on instance content \citep{ref27,ref36}. These contrasts target different contributions: an approximate budget allowance does not match realized resource use, and precomputed candidate outcomes do not capture feedback under a changed online policy.

Executing a changed policy can affect which logged states remain valid for comparison. Replay Gap finds low logged-state validity after model switching in two quantized configurations within one low-success pilot \citep{ref37}. This identifies a limitation of the tested log-splicing procedure. Executing the changed policy and estimating variation across reruns under the same conditions would help distinguish an assignment advantage from invalid state reuse or ordinary execution variability.

The second application asks whether \textbf{First-chunk selection} supports changing a coding agent's ordering policy to improve issue resolution. The intervention replaces FIFO with prioritization on tasks selected for gold-file reachability. The measured endpoints are placement quality and offline localization; the paired localization tests are inconclusive. The supported conclusion concerns placement and localization on this subset, without establishing equivalence or A for issue resolution. The missing comparison executes both ordering policies through issue resolution on the same tasks, with the executor and resource allowance retained. Attribution to placement would additionally require controlling context exposure \citep{ref26}.

The third application asks whether \textbf{Evidence-Carrying Termination} supports a safeguard that reduces unsupported stopping while preserving completion. The intervention changes the stopping rule while retaining the planner configuration, prompt, tools, and nominal checkpoint, with independent planner calls in each arm. Its closed-loop confirmation reports completion of 97/132 versus 92/132, a +3.79-point difference with a task-cluster interval of [0.00, 9.09] across three seeds (source section ``Closed-Loop Confirmation'', source Table 4). This supports non-inferiority under the prespecified \ensuremath{-}10-point margin, while completion superiority remains unestablished. Premature unsupported termination falls from 40/66 to 0/66, a separate positive endpoint. The supported engineering use is the safeguard under those tested conditions. If the objective is B, the next comparison must connect quality changes to completion while retaining the response interface; alignment of endpoints with distinct denominators does not supply that attribution. The deterministic tools and substitute model also delimit the setting in which the safeguard was tested \citep{ref29}.

We interpret positive and inconclusive results using the same criteria: first identify the replacement and the endpoint measured during execution, then examine comparability, controls, and uncertainty. E2 records an executed evaluation; it does not establish attribution to decision quality. Likewise, a non-significant test is not evidence of equality, as illustrated by the inconclusive comparison in Scores Are Not Decisions \citep{ref38}. Evidence can establish the value of a replacement within a specific setting while leaving its mechanism unresolved; a useful engineering result need not resolve every mechanism.

\subsection{Eight reporting items, conditional on the claim}

Table 5 links the comparison analysis to reporting choices. A concerns task improvement from the specified replacement; B concerns local-quality attribution. Performance preservation under a non-inferiority or equivalence criterion is a separate claim. Trigger, transfer, and efficiency items apply when those claims are made. The table provides guidance rather than a validated scoring instrument; the eight items are not universal prerequisites. A does not always require a local-quality metric, whereas claims of maintained performance require an appropriate non-inferiority or equivalence design.

\begin{table}[!htbp]
\caption{Conditional reporting guide: the claim determines the needed comparison. Items are proposed guidance, not a validated quality score.}
\centering
\footnotesize\setlength{\tabcolsep}{3pt}
\begin{tabular}{@{}>{\raggedright\arraybackslash}p{0.18\textwidth}>{\raggedright\arraybackslash}p{0.25\textwidth}>{\raggedright\arraybackslash}p{0.25\textwidth}>{\raggedright\arraybackslash}p{0.25\textwidth}@{}}
\toprule
Reporting item & Applicable claim & Comparison or information needed & Case motivating the item \\
\midrule
1. Executed comparison and endpoint & A or performance preservation; distinguish proxy or offline scope & Name the replacement, comparator, task population, endpoint, and fixed factors; execute the changed policy for a closed-loop task claim. & First-chunk localization cannot establish issue-resolution benefit; ECT measures executed completion. \\
\addlinespace[4pt]
2. Recovery and disruption & B when an intervention can rescue or harm trajectories & Report paired rescued failures and disrupted successes, with their baseline denominators; distinguish prediction quality from action advantage. & Failure Prediction and Calibration separate risk from intervention value. \\
\addlinespace[4pt]
3. Trigger-conditioned effects & Claims about triggered versus untriggered tasks & Use a common fixed task partition and report both groups; when treatment changes triggers, use same-prefix branches or a design addressing selection. & Outcome Monitors' always-warn arm also warns where the detector is silent; BoR's choice accuracy is conditional on presentation; Calibration supplies same-prefix branches. \\
\addlinespace[4pt]
4. Controls matched to attribution & A at package scope; B for an identified component & Retain the response policy, relevant information, and budget; match choice quota when the claim concerns selection at that quota. Combine effects only across aligned conditions. & Outcome Monitors' timing placebo fixes envelope and count, while its package and receipt contrasts use different policies and populations; HYSET retains architecture differences. \\
\addlinespace[4pt]
5. Uncertainty and reruns & A or performance preservation; additional checks for assignment-based B & Report repetitions and intervals or paired tests appropriate to the design. Estimate variation across same-condition reruns before interpreting oracle headroom or assignment gains. & ECT specifies a non-inferiority margin; First-chunk's non-significance is not equivalence. \\
\addlinespace[4pt]
6. Measurement and information visibility & B; feasibility of deployable A & Define labels and denominators, connect them to the executed contrast, and distinguish available inputs from future outcomes or oracle labels. & TabAgent reference-path labels do not establish necessity; AgentRouter's true-quality fallback bounds deployment claims. \\
\addlinespace[4pt]
7. Transfer conditions & Claims of transfer across backbones or environments & State which rule, threshold, and inputs stay fixed; report retraining or recalibration separately from cold transfer. & SWE-PRM and later configurations differ; multiple-backbone testing alone does not isolate fixed-rule transfer. \\
\addlinespace[4pt]
8. Complete cost accounting & Efficiency or budget claims & Include controller, retry, fallback, and judging costs; distinguish budget allowances from realized use and identify relevant hardware or price dates. & The supplementary budget comparison uses an approximate allowance, with unequal realized token use. \\
\addlinespace[4pt]
\bottomrule
\end{tabular}
\end{table}

Items 6 and 7 address the meaning of a local score and the availability of its required inputs during deployment. TabAgent labels tools used in successful reference trajectories; these labels need not identify necessary actions or cover all useful alternatives. AgentRouter evaluates a fallback with true quality information, while its deployment description uses a proxy whose same-task online controller performance was not independently verified \citep{ref39,ref40}. Calibration Is Not Control likewise distinguishes predicting failure from identifying a beneficial intervention \citep{ref28}. These examples show why interpreting a reported local improvement requires attention to label meaning and input availability.

For item 3, under a common fixed partition T, the overall effect is q\ensuremath{\Delta}T + (1 \ensuremath{-} q)\ensuremath{\Delta}notT, where q is the fraction in T. Omitting the second term needs justification; treatment-dependent triggering does not create comparable groups automatically. For item 5, baseline reruns are a proposed safeguard against attributing stochastic variation to assignment; the selected studies do not validate a universal minimum number of runs.

Reporting choices depend on the claim rather than the decision category alone. Monitoring and gating often call for recovery and trigger analysis; routing may require information and allocation controls; selection may require measuring how a shortlist is consumed. The record base for these suggestions differs by category (Table 3): 37 selection and 13 monitoring records, but four routing records and none for assistance or budgeting, so the routing suggestion rests mainly on the cited studies' own designs. The ten decision categories organize the studies; they do not imply that every study in a category requires the same evaluation protocol.

\subsection{Consequences for agent design}

Adopting a complete controller and replacing one of its internal components are different engineering decisions. A positive executed package comparison can justify retaining that configuration for the tested setting, even while its mechanism remains unresolved. Optimizing the detector alone, changing the recovery interface, or transferring the rule to a new executor asks whether the benefit survives a different combination. The relevant evidence is a comparison that retains the partner component or measures the interaction, as specified in Section 2.5. A recovery action list, retained task state, or executor context belongs in that specification whenever the replacement changes it.

The same distinction applies to choosing an evaluation endpoint. Offline placement or localization can guide component development, but deciding whether to deploy it for issue resolution requires executing that task. A reduction in unsupported stopping can justify a termination safeguard when completion is preserved under the stated margin; it does not establish a completion improvement. These choices retain the useful results in Table 4 while connecting each to the particular decision it supports. The next evaluation should measure the unresolved outcome or vary the unresolved component, rather than repeat a broader package comparison that leaves the same question open.

\subsection{Research opportunities}

\textbf{Separate warning quality from recovery capacity.} A prospective comparison could cross two warning rules with two response policies on a common task population, holding the executor, allowed inputs, and budget allowance fixed. This supplies the four conditions needed to estimate whether the rule effect changes with the response policy. For monitors, the more informative rule contrast is selection at a fixed quota rather than timing, which Outcome Monitors' placebo has already examined. A timing comparison could then vary the intervention checkpoint and state-retention policy, measuring rescued failures, disrupted successes, completion, and remaining cost. These designs would test the recovery and timing explanations without combining unmatched contrasts or treating differences between Last Step Matters and RestartSmart as a causal experiment.

\textbf{Compare static, replay, and executed rankings for the same decision.} Hold the candidate policies, task distribution, and budget definition fixed; vary the evaluation mode; and measure ranking uncertainty, the validity of reused states, and task completion. Replay Gap identifies a concrete failure mode of state reuse. LoopArena reports a positively correlated slice/full-task ordering of six controllers under its main Core criterion (Spearman \ensuremath{\rho} = 0.74; two of thirteen strictly ordered pairs reverse), while TwinRouterBench provides static and live tracks; these observations motivate a controlled comparison rather than establish agreement under every scoring rule or for every router \citep{ref41,ref42}. Independent reruns would help separate a change in ranking from ordinary execution variability.

\textbf{Test information and consumption as competing explanations.} Hold the executor, available tools, intervention content, and budget fixed; vary the selector or deployment-visible quality signal; and record what was selected, what was actually consumed, and the task endpoint. This comparison addresses the gaps identified in HYSET, First-chunk, TabAgent, and AgentRouter. A condition with oracle information can estimate headroom, but a separate deployable condition is needed to estimate achievable task benefit (A). Attribution to local decision quality (B) then requires matched local measurements and controls that retain the other components.

\textbf{Separate cold transfer from adaptation.} Freeze a controller, threshold, label definition, and allowed inputs before changing the backbone or environment; report that result separately from a recalibrated condition. Measure local decision quality, recovery and disruption where applicable, completion, and total cost. Track whether any change reflects a mismatch in label meaning, unavailable inputs, or altered recovery capacity. Tests on several backbones do not by themselves establish that one unchanged controller transfers between them. The differing directions of process-reward intervention results in SWE-PRM and later agent configurations motivate this distinction, but the differences between configurations do not isolate a causal property of the model \citep{ref43,ref25}. These are proposed comparisons derived from the evidence gaps; this review does not execute them.

\section{Limitations}

The evidence base is recent and largely unrefereed: 272 of the 348 included studies (78\%) are preprint or repository records, and 269 are dated 2026. Reported results may change as these studies are revised or formally published.

Language models assisted screening, extraction, adjudication, and the record-level annotations (Section 2.7). The agreement statistics compare model-assisted coding streams, and the record annotations are single-pass judgments without human double coding. The source reconciliation in this revision corrected several annotations and extraction fields, which shows that individual judgments can be wrong; the counts in Tables 2 and 3 carry that uncertainty.

The comparison records are a purposive collection. Some were added to check claims in this manuscript, and no assistance or budgeting study contributed a record, so differences between categories reflect what was extracted as well as what was studied. The worked cases illustrate the procedure rather than validate it.

\section{Conclusion}

The five-step analysis connects a reported comparison to the decision it can support. Applied to the retained records, it distinguishes executed task evidence from restricted endpoints and diagnostics, and shows that the evidence for attribution is narrower than the evidence for adoption: fewer than half of the task contrasts report contrast-level uncertainty, and only nine hold the response policy fixed. Within the task comparisons, rule, response and package changes answer different attribution questions. The worked cases preserve useful conclusions: a beneficial monitoring package can be retained while its detector selection remains untested, an improved placement metric can motivate an executed task comparison, and a termination safeguard can reduce unsupported stops while preserving completion under a specified margin. The resulting record-level synthesis and conditional reporting guide specify which conditions must be retained, which outcome remains to be measured, and when existing component effects cannot be combined. They provide an inspectable basis for choosing the next comparison without discarding benefits already established by the source evidence.

\section*{Data and code availability}
The study-level coding, comparison records, included-study catalogue, search and screening records, and code for reproducing the descriptive summaries and agreement statistics are provided in a supplementary archive that accompanies the journal submission and is available from the corresponding author on request. It also includes the interpretation codebook, every record-level judgment, the log of corrections with historical values, and code for reproducing the synthesis counts and tables. Source publications are identified through bibliographic references and persistent links. The supplement does not redistribute third-party full texts or database abstracts. The data and code in the supplement are licensed under CC BY 4.0.
\begin{acks}
No funding was received for this work. The authors have no relevant financial or non-financial interests to disclose. Shuyang Zhang and Jianshuo Chang contributed equally to all stages of this review, including conceptualization, methodology, literature retrieval and screening, coding, evidence synthesis, analysis, and manuscript preparation. The authors performed their coding work separately.
\end{acks}
\bibliographystyle{ACM-Reference-Format}
\bibliography{refs}
\appendix
\section{Search, coding, and sensitivity details}

\subsection{Search and selection records}

The arXiv and supplementary searches were performed on 24 September 2026. The first arXiv query combined agent, language-model, control-action, and outcome blocks and was restricted to computer-science categories and submission dates from 1 September 2023 to 24 September 2026. It returned 4,656 records. Title screening and a keyword recheck produced 310 records for abstract screening. Final decisions after source-based eligibility checks were 134 included, 84 background, and 92 excluded; original screening decisions and subsequent corrections are retained in the records.

An expanded arXiv query added control-action terms and removed the outcome block. It returned 15,292 records, including 10,518 new records after deduplication against previously seen records across the search arms. Abstract screening covered 292 records: 125 included, 92 background, and 75 excluded, including two historical adjudication exclusions. Recall against already known eligible arXiv studies was 145/146. Because the search vocabulary was developed using known missed studies, this check does not independently estimate search sensitivity.

OpenAlex, Semantic Scholar citation tracking, ACL Anthology, and OpenReview contributed 11, 11, zero, and two additional studies, respectively. The ACL search used the official bibliography with abstracts dated 22 September 2026. OpenReview searches used 139 query groups; 68 reached the first-page result limit and may therefore have been truncated. Exact fields, queries, deduplication rules, and arm-level logs accompany the review.

Scopus was searched through institutional access on 25 September 2026 using title/abstract fields, the computer-science subject restriction, and publication year later than 2022. The export contained 2,985 records. Removing 199 non-primary document types, 54 previously seen records, and three internal duplicates left 2,729 new records. Title screening retained 593, and keyword rechecking added three. Among the 596 abstracts, the initial model-assisted screen proposed 305 for inclusion. The log records author review of this subset, resulting in 65 included, 58 background, 177 excluded, and five duplicates. The other 291 records retained their initial classifications: 119 background and 172 excluded. Consequently, the complete abstract-stage ledger contains 65 included, 177 background, 349 excluded, and five duplicates. A random recheck of 50 initially unselected records found no additional eligible study; this does not establish perfect recall.

The Scopus query notes describe a September 2023 cutoff applied after export, but the available export records only the year, so the month-level cutoff could not be verified. We therefore report the executable year restriction without claiming that the month filter was fully implemented. DBLP access was blocked by a human-verification challenge. Web of Science, ACM Digital Library, and IEEE Xplore were not separately searched. Scopus coverage is not assumed to be identical to direct searches of those resources.

Four coding tables contain 348 unique normalized study identifiers: 134 first-round arXiv, 125 second-round arXiv, 24 supplementary, and 65 Scopus studies. For the arXiv and Scopus searches, record-level inclusion sets reconcile with the totals at each stage. Some supplementary search stages are documented only in logs, limiting verification at the record level.

\subsection{Study-level coding rules}

We coded ten decision points: selection or shortlisting (SEL), invocation necessity or skipping (NEC), step-level routing or escalation (RTE), termination or restart (TRM), monitoring followed by intervention (MON), pre-execution gating (GATE), requesting assistance (ASK), computational budgeting (BUD), deterministic replacement (REP), and task-level configuration (CFG). Controller types include rules or algorithms, classifiers, probes, language-model judges, smaller fine-tuned models, retrievers, and reinforcement-learning or bandit policies.

Four evaluation fields describe each study. A local metric quantifies the replaced decision's quality; resource use and post-intervention recovery do not automatically qualify. Task endpoints are measured success, pass, or completion (Y), proxies such as cost, substep accuracy, or probability-based simulated success (P), or no reported endpoint (N). Evaluation levels are static/offline prediction (E0), replay, precomputed outcomes, simulated execution, or offline evaluation over fixed prefixes (E1), and online closed-loop comparison on the same task set (E2). Executing branches to collect outcomes does not by itself qualify a controller comparison as E2; the compared decision policy must be evaluated in the closed loop. Subsequent extraction assigned the level supporting the main claim. Multiple-backbone coding records tests on different agent backbones; it does not establish transfer of a fixed controller. F denotes insufficient information.

The \texttt{cross\_model\_env} field records tests on multiple agent backbones. Changes to the benchmark, controller size, or model roles alone do not qualify. Current codes are used for the descriptive summaries; original values and reasons for corrections remain in the provenance records.

\subsection{Verification, adjudication, and agreement}

\textbf{Verification coverage.} Source-specific records document checks for 140 studies at varying depths, including some reassessments based on excerpts or retained notes. For the other 208 studies, two model-assisted coding streams were assigned to extract the four evaluation fields. Full-text extraction records cover the entire group; the agreement analysis includes only valid pairs of outputs. Source-linked checks also addressed ambiguous decision-point and controller assignments. These checks did not constitute a complete re-adjudication of eligibility.

\textbf{Adjudication.} Disputed fields were resolved with model assistance under the recorded version-2 rules. Source mismatches were corrected using verified documents, and source-specific decisions were recorded separately from the original paired outputs. The current tables incorporate the resulting codes. The provenance records retain earlier values, source locations, and correction reasons.

\textbf{Retrospective rule check.} A model-assisted check screened 238 earlier records against the recorded version-2 definitions. Potential differences in 155 studies were reviewed against source texts and prior evidence. This review led to 108 field corrections across 85 studies. One field remained unresolved and retained its earlier value. A separate correction ledger updates the current descriptive tables while preserving the original extraction outputs and agreement calculations.

\textbf{Agreement sample.} Stream 2 used the earlier instructions for its first 100 assignments and version 2 for the next 108; the prompt version used by stream 1 could not be fully established. The extraction instructions required information isolation, but adherence and the identity of every historical operator could not be independently confirmed. Agreement therefore describes the coding outputs received, rather than independent human double coding under uniform conditions. Missing outputs, source mismatches, and invalid values are excluded; disagreements related to coding rules are retained. Re-extraction from corrected sources does not retroactively create independent pairs.

\begin{table}[!htbp]
\caption{Pre-adjudication agreement of received coding streams, separated by assignment batch.}
\centering
\small
\begin{tabular}{llrrr}
\toprule
Original-output scope & Field & Valid pairs & Agreement & Kappa \\
\midrule
All & Local metric & 196 & 0.903 & 0.799 \\
All & Task endpoint & 197 & 0.914 & 0.809 \\
All & Evidence level & 197 & 0.827 & 0.670 \\
All & Multiple backbones & 197 & 0.858 & 0.734 \\
First 100 & Local metric & 99 & 0.899 & 0.783 \\
First 100 & Task endpoint & 100 & 0.900 & 0.792 \\
First 100 & Evidence level & 100 & 0.810 & 0.649 \\
First 100 & Multiple backbones & 100 & 0.840 & 0.698 \\
Next 108 & Local metric & 97 & 0.907 & 0.812 \\
Next 108 & Task endpoint & 97 & 0.928 & 0.829 \\
Next 108 & Evidence level & 97 & 0.845 & 0.680 \\
Next 108 & Multiple backbones & 97 & 0.876 & 0.758 \\
\bottomrule
\end{tabular}
\end{table}

\textbf{Changes from initial codes.} In the 200-study coding-update audit, adjudicated values differed from the initial abstract-based codes in 80 local-metric fields (40.0\%), 60 task endpoints (30.0\%), 36 evidence levels (18.0\%), and 71 multiple-backbone fields (35.5\%). These figures describe the recorded adjudication update; source-specific corrections are tracked separately. Because coding-rule changes and adjudication accompanied the more detailed reading, these percentages are neither corpus-wide error rates for abstract-based coding nor isolated effects of full-text access. At least one evaluation field changed in 148 studies during that adjudication update.

The current coding tables contain 16 cells marked as insufficient information across the four evaluation fields. Agreement statistics and update counts describe their stated samples; they are separate from the coverage of completed source checks.

\subsection{Comparison-level extraction and case selection}

The mechanism synthesis operates at the comparison level rather than assigning a single local--task relationship to an entire paper. Each structured record preserves the component, A/B conditions, backbone, task subset, measurements, denominators, repetitions, uncertainty, tests, and source locations when reported. Missing endpoint-type and evaluation-level codes have an explicit status: not extracted, not reported, or not applicable; a populated historical code is marked recorded. A not-extracted status does not mean that the information is absent from the source, and study-level codes are not copied into individual comparisons. Historical M/J endpoint codes retain their measured/LLM-judged meanings and are not recoded into the study-level Y/P/N scheme. Original measurements are preserved in neutral reported-metric fields. Only measurements classified as decision quality populate the local-quality fields; resource use, intervention frequency, downstream outcomes, and evaluation diagnostics remain separate. A comparison without a local baseline can document coexistence of levels but cannot establish a local improvement. A proxy endpoint remains distinct from task completion. A non-significant difference is not coded as equality or non-inferiority.

The comparison collection contains 94 records from 42 studies; 92 records from 40 studies correspond to entries in the 348-study inclusion set. The other two are supplementary evidence: When Better Turns Do Not Make Better Agents was screened as background \citep{ref44}, and the budget-constrained study of online skill and memory modules appears among the citation-tracking candidates but is absent from the four inclusion tables. The former compares gold-history and closed-loop evaluation of the same fine-tuned models and is read as a diagnostic, because fine-tuning is outside the replacement scope; the latter supplies a budget-allocation control for interpreting system gains. Neither changes the inclusion denominator. Every record identifies its membership and selection rationale. The collection includes comparisons of checkpoint recovery and retrieval efficiency; the eight cases in Table 4 provide the main worked examples.

Cases were selected purposively from retained comparison extractions and records needed to check manuscript claims. The eight worked examples cover the three candidate mechanisms, positive net-benefit evidence, attribution and deployment questions, and proxy or diagnostic endpoints. Seven are included studies and one is supplementary. This selection supports the interpretation of specific comparisons; it neither exhaustively covers the 191 co-reporting studies nor provides a basis for estimating prevalence. Source-location and semantic checks are retained with each comparison.

The assessment applies the following precedence to each record. If arm identities, endpoint identity, or essential paired results cannot be reconciled from the extract, the record is first checked against its source. Then a cross-setting range, stage profile, scoring change, or single-method description without a common component replacement is a diagnostic summary. An identifiable comparison with only a component, intermediate, offline/replayed, probability-simulation, or mixed non-agent endpoint receives a restricted scope. A comparison of executed policies with a task outcome receives the task-contrast scope, including judged outcomes and synthetic tool environments where applicable. Missing uncertainty alone does not remove an otherwise identifiable task contrast; its conclusion remains a point estimate. The scopes are assigned to records, not entire source papers, and distinct-study counts can overlap across scopes.

Each annotation separately identifies the apparent intervention object, evidence for retained response conditions, supported reading, documented design boundary, information gap, and next comparison. The original measurements and historical evaluation codes remain available alongside this interpretive layer. Endpoint definitions take precedence when historical labels and inspected protocols differ. The supplementary files provide the full annotation table, source locators, targeted-check notes, and count-reproduction code. This pass supplies no new intercoder-agreement estimate.

In the October 2026 revision, every record that the earlier pass had left unresolved was read against its source. All nine were reconciled: three became task contrasts, five restricted comparisons, and one a diagnostic. Two further records were reclassified as diagnostics for consistency with the rules above, and corrections to arms, endpoints, denominators, and uncertainty were made in 12 other records. The full-text source for TabAgent was obtained for this check. Four partner records were added because a retained source already reported them: the Outcome Monitors timing placebo and always-warn contrasts, and the two First-chunk models previously omitted from the extract. Every change is logged with its historical value, revised value, reason, and source locator. Detailed case prose can still examine an additional source contrast, such as the learned-versus-schema-only detector ablation, without adding it to the record set.

\subsection{Descriptive summaries and sensitivity}

The accompanying records provide overall and stratified summaries by search source and verification history. Pooled frequencies summarize the current coding, but differences in rule versions and checking depth mean that the strata do not reflect a uniform measurement process. Bibliographic venue labels and future volume years do not establish publication by the search date, so we do not report an unverified count of formally published studies.

Excluding 11 specifically identified scope-boundary studies leaves 337 studies, with 217 local-Y, 139 local/measured co-reporting, 47 local/proxy co-reporting, and 223 E2 codes. A majority still reported a local metric. This purposive sensitivity analysis is not a complete eligibility audit.

\section{Included-study category index}

The following category index identifies every included record by its primary decision point. Study-level coding defines membership in this index; the comparison-level analyses in Sections 3.4--3.5 assess the selected findings and their limits.

\textbf{Selection and shortlisting (SEL; 84 studies).} \citep{ref45,ref20,ref39,ref46,ref47,ref48,ref49,ref50,ref51,ref52,ref53,ref54,ref18,ref38,ref55,ref56,ref57,ref58,ref59,ref19,ref60,ref26,ref61,ref62,ref63,ref21,ref64,ref17,ref16,ref65,ref66,ref67,ref68,ref69,ref14,ref70,ref71,ref72,ref73,ref74,ref15,ref75,ref23,ref76,ref77,ref78,ref79,ref80,ref81,ref82,ref83,ref84,ref85,ref86,ref87,ref88,ref89,ref90,ref91,ref92,ref93,ref94,ref95,ref96,ref97,ref98,ref99,ref100,ref101,ref102,ref35,ref103,ref104,ref105,ref106,ref107,ref108,ref109,ref110,ref111,ref112,ref113,ref114,ref115}.

\textbf{Routing (RTE; 56 studies).} \citep{ref116,ref117,ref118,ref119,ref120,ref121,ref122,ref123,ref124,ref125,ref126,ref127,ref128,ref129,ref130,ref131,ref132,ref133,ref134,ref135,ref136,ref137,ref138,ref36,ref139,ref140,ref141,ref40,ref142,ref143,ref37,ref144,ref145,ref146,ref147,ref148,ref149,ref150,ref151,ref152,ref153,ref154,ref155,ref156,ref157,ref158,ref159,ref160,ref161,ref162,ref163,ref164,ref165,ref166,ref167,ref42}.

\textbf{Monitoring and intervention (MON; 49 studies).} \citep{ref168,ref169,ref24,ref170,ref171,ref172,ref173,ref174,ref175,ref176,ref177,ref178,ref179,ref180,ref181,ref182,ref183,ref184,ref185,ref186,ref187,ref188,ref189,ref190,ref191,ref192,ref193,ref194,ref195,ref196,ref197,ref28,ref43,ref1,ref198,ref199,ref200,ref201,ref32,ref202,ref203,ref204,ref205,ref206,ref207,ref208,ref209,ref210,ref41}.

\textbf{Gating (GATE; 41 studies).} \citep{ref30,ref211,ref212,ref213,ref214,ref215,ref216,ref217,ref218,ref219,ref220,ref29,ref31,ref221,ref222,ref223,ref224,ref225,ref226,ref227,ref228,ref229,ref230,ref231,ref232,ref233,ref234,ref235,ref236,ref237,ref238,ref239,ref240,ref241,ref242,ref243,ref244,ref245,ref246,ref247,ref248}.

\textbf{Termination and restart (TRM; 32 studies).} \citep{ref249,ref22,ref250,ref251,ref252,ref253,ref254,ref255,ref256,ref25,ref33,ref257,ref258,ref259,ref260,ref261,ref262,ref263,ref264,ref265,ref266,ref267,ref268,ref269,ref270,ref271,ref272,ref273,ref274,ref275,ref276,ref277}.

\textbf{Task configuration (CFG; 27 studies).} \citep{ref13,ref278,ref279,ref280,ref281,ref282,ref283,ref284,ref285,ref286,ref287,ref288,ref289,ref290,ref291,ref292,ref293,ref294,ref295,ref296,ref297,ref298,ref299,ref300,ref301,ref302,ref303}.

\textbf{Necessity decisions (NEC; 24 studies).} \citep{ref304,ref305,ref306,ref10,ref307,ref308,ref309,ref310,ref311,ref312,ref313,ref314,ref34,ref315,ref316,ref317,ref318,ref319,ref320,ref321,ref322,ref323,ref324,ref325}.

\textbf{Assistance decisions (ASK; 14 studies).} \citep{ref9,ref326,ref327,ref328,ref329,ref330,ref331,ref332,ref333,ref334,ref335,ref336,ref337,ref338}.

\textbf{Budgeting (BUD; 12 studies).} \citep{ref11,ref339,ref340,ref341,ref342,ref343,ref344,ref345,ref346,ref347,ref348,ref349}.

\textbf{Deterministic replacement (REP; 9 studies).} \citep{ref350,ref351,ref352,ref353,ref12,ref354,ref355,ref356,ref357}.

\end{document}